\documentclass[conference]{IEEEtran}
\IEEEoverridecommandlockouts
\usepackage{cite}
\usepackage{amsmath,amssymb,amsfonts}
\usepackage{algorithmic}
\usepackage{graphicx}
\usepackage{textcomp, tabularray, setspace, subcaption, adjustbox, amsmath, amsfonts}
\usepackage{fancyhdr}

\usepackage{rotating}
\usepackage{multirow}

\usepackage{graphicx} 
\usepackage{wrapfig}
\usepackage{enumerate}
\usepackage{hyperref}
\usepackage{xurl}
\usepackage{tabularx}
\usepackage{url}
\usepackage{dcolumn}
\usepackage{listings}
\usepackage{algorithm}
\usepackage{amsmath}
\usepackage{verbatim}
\usepackage{float}
\usepackage{graphicx}

\usepackage[table,xcdraw]{xcolor}

\definecolor{headercolor}{rgb}{0.2,0.4,0.7}
\definecolor{rowcolor}{rgb}{0.9,0.9,0.9}
\definecolor{textcolor}{rgb}{0.9,0.9,0.9} 

\title{Correlation of Object Detection Performance with Visual Saliency and Depth Estimation}

\author{
    \IEEEauthorblockN{Matthias Bartolo}
    \IEEEauthorblockA{\textit{Dept. of Artificial Intelligence}  \\ \textit{University of Malta} \\
    \href{mailto:matthias.bartolo@ieee.org}{matthias.bartolo@ieee.org}}

    \and
    \IEEEauthorblockN{Dylan Seychell}
    \IEEEauthorblockA{\textit{Dept. of Artificial Intelligence}  \\ \textit{University of Malta}  \\
    \href{mailto:dylan.seychell@ieee.org}{dylan.seychell@ieee.org}}
}

\date{30th September 2024}

\begin{document}

\maketitle
\thispagestyle{plain}
\pagestyle{plain}

\begin{abstract}

As object detection techniques continue to evolve, understanding their relationships with complementary visual tasks becomes crucial for optimising model architectures and computational resources. This paper investigates the correlations between object detection accuracy and two fundamental visual tasks: depth prediction and visual saliency prediction. Through comprehensive experiments using state-of-the-art models (DeepGaze IIE, Depth Anything, DPT-Large, and Itti's model) on COCO and Pascal VOC datasets, we find that visual saliency shows consistently stronger correlations with object detection accuracy (mA$\rho$ up to 0.459 on Pascal VOC) compared to depth prediction (mA$\rho$ up to 0.283). Our analysis reveals significant variations in these correlations across object categories, with larger objects showing correlation values up to three times higher than smaller objects. These findings suggest incorporating visual saliency features into object detection architectures could be more beneficial than depth information, particularly for specific object categories. The observed category-specific variations also provide insights for targeted feature engineering and dataset design improvements, potentially leading to more efficient and accurate object detection systems.
\end{abstract}



\section{Introduction}%
\label{chp:intro}

Despite the progress being made in computer vision, object detection remains a fundamental challenge, with current approaches achieving impressive accuracy but still facing limitations in complex scenarios \cite{human_perception, bartolo2024integrating}. Human perception is influenced by various factors, leading to selective attention towards certain elements in our environment.  The same trait is followed in machines, which also rely on mechanisms that prioritise specific aspects of images \cite{perception}. Whilst advances in deep learning have led to significant improvements, understanding how different visual tasks relate to and potentially enhance object detection performance remains crucial for further progress. The relationships between complementary tasks such as depth estimation and visual saliency prediction could provide valuable insights for improving detection systems.

Recent works in computer vision have explored multi-task learning approaches, combining object detection with either depth estimation or saliency prediction \cite{vandenhende2022multitasklearningvisualscene}. However, these studies typically focus on end-to-end performance rather than analysing the underlying correlations between these tasks. Understanding these correlations provides opportunities for advancement.  These include improvement of object detection architectures while also improving computational efficiency in the process.  Moreover, it also informs researchers on how datasets can be designed to improve these computer vision tasks. The challenges of detecting objects across varying scales and contexts \cite{Pisani2024DetectingLF, small_detection} make it essential to understand which complementary tasks provide the most beneficial information for object detection. Additionally, such advancements are also important when looking at AI techniques from a sustainable perspective.  

This paper presents an investigation of the relationships between depth estimation and visual saliency. We present an examination of how each factor correlates with object detection performance. Through analysis across different datasets and models, we aim to explore whether significant correlations exist between these visual tasks and object detection accuracy. This investigation focuses particularly on how these correlations vary across different object categories and scales since it provides insights that could inform more effective object detection architectures.

The implications of the work presented in this paper extend beyond a theoretical understanding of computer vision.  They present practical perspectives for refining object detection systems. The quantification of correlations between these tasks provides empirical evidence for which auxiliary features might be most beneficial for enhancing detection performance. Furthermore, our analysis of category-specific variations offers insights for targeted improvements in both model architecture and dataset design.

\section{Background}%
\label{chp:background}

\begin{figure*}[t]
    \centering
    \includegraphics[width=\linewidth]{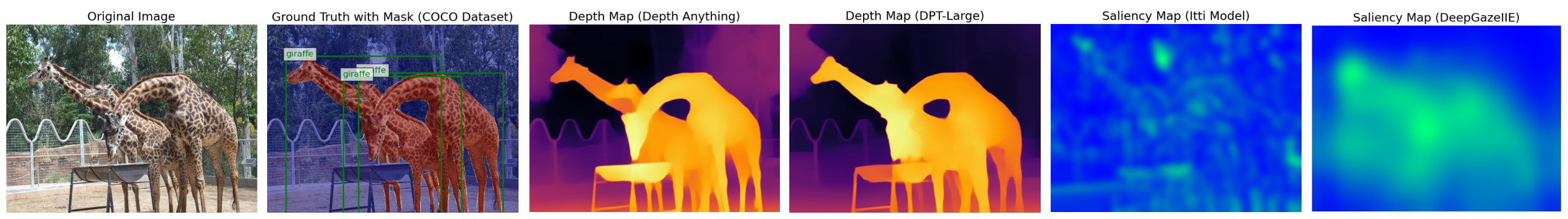}
    \caption{Comparison of outputs generated from various saliency and depth prediction models alongside the original image and annotations.}
    \label{fig:image_compare}
\end{figure*}

\subsection{Object Detection}
\label{sec:detection}

Within the field of computer vision, object detection can be seen as a critical problem in recognising and localising objects inside various images and videos. While the task may seem straightforward at first, its difficulty arises from the immense variability in object size, shape, orientation, occlusion, and lighting. Moreover, the context in which these objects appear adds another layer of complexity, requiring AI models capable of generalising across diverse environments and perspectives. Although notable progress has been made in this area, with state-of-the-art technologies such as YOLOv10 \cite{yolov10}, YOLO11 \cite{yolo11} and RT-DETR \cite{rtdetr}, utilising advanced algorithms to tackle these challenges, achieving human-level proficiency remains a significant challenge. Understanding the intricacies of human visual perception \cite{human_perception}, particularly how individuals intuitively detect and locate objects, is key to further advancing object detection systems.

\subsection{Depth Prediction}
\label{sec:depth}

Conversely, depth prediction involves determining the distance of each pixel in an image relative to the camera, effectively reconstructing a scene in three dimensions \cite{depth_estimation}. This can be achieved using either monocular images (from a single viewpoint) or stereo images (from multiple perspectives of the same scene). Traditional approaches rely on multi-view geometry to establish the spatial relationships between images and calculate depth. However, recent advances in deep learning have introduced more sophisticated techniques \cite{depth_anything}, allowing for more accurate and robust depth prediction from even single images, bypassing the limitations of classical geometric methods. These modern approaches have proven crucial in applications such as autonomous driving, robotics, and augmented reality.

\subsection{Saliency Prediction}
\label{sec:saliency}

Visual saliency refers to the ability to identify regions in an image that are most likely to attract human attention or be important for machine learning models \cite{itti}. Saliency maps are used to highlight these areas, showing where a viewer's gaze naturally lands or which parts of an image hold the greatest relevance for computational analysis. While early approaches such as \cite{itti} relied on basic visual features such as contrast, color, and edges, advances in deep learning have enabled more accurate predictions \cite{deepgaze}. However, saliency is inherently subjective, varying with individual perception, context, and the task at hand, making its prediction a complex challenge \cite{sara}.

\section{Related Work}%
\label{chp:literature}

Studies exploring the correlation between object location and various AI techniques have been conducted to understand how individuals perceive and locate objects. In particular, T. Boger and T. Ullman \cite{object_detection_philospy} performed a series of tests to examine how people determine the position of objects. Their experimental setup involved 50 participants, each tasked with clicking on the centre of mass for 50 randomly assigned images. The authors also evaluated eight AI models to assess the correlation between participant input and model predictions, assessing the models' adaptability and how closely their performance aligns with human-like proficiency.

By varying the stimuli in each experiment, they found that physical reasoning, specifically using the centre of mass, consistently plays the most significant role in perceived object location, regardless of the object's realism. However, while insightful, this study is limited by the small number of images and its focus solely on static objects, leaving the perception of dynamic objects in non-iconic scenes unexplored.

\section{Methodology}%
\label{chp:methodology}

This paper proposes a series of experiments to evaluate the correlation between depth prediction and saliency prediction in object detection on a larger scale using popular object detection datasets. These experiments are designed to test how well depth and saliency prediction techniques align with object detection performance. 

\subsection{Datasets}%
\label{sec:datasets}

Two popular object detection datasets with similar structures were used as the basis of the experimental structure to evaluate these approaches. The widely used COCO dataset \cite{cocodataset} was chosen to assess the correlation across a diverse range of images, featuring 80 categories and over 200,000 images. The smaller and less commonly used Pascal VOC dataset \cite{pascal_voc_2012} was also included to examine the setup on a dataset with 20 categories and over 11,000 images that feature larger objects. Both datasets include segmentation masks that serve as ground truth for evaluating the performance of the prediction models. The experimental setup was tested on the COCO 2017 training dataset and the Pascal VOC 2012 dataset.

\subsection{Models}%
\label{sec:models}

To explore the correlation between object detection, visual saliency, and depth prediction, four prediction models were utilised: two depth prediction models and two saliency prediction models.

\textbf{Depth Anything} -- A widely recognised depth prediction network that is known for its robustness in monocular depth prediction across various conditions. It serves as a foundational model tested on numerous public datasets \cite{depth_anything}.

\textbf{DPT-Large} -- Was selected as another depth prediction model due to its advanced architecture, which incorporates dense vision transformers. Despite its large size, DPT-Large is effective for monocular depth prediction and includes techniques to address the loss of feature granularity \cite{dpt_large}.

\textbf{Itti's Visual Attention Model} -- Inspired by the behaviour and neurology of the early primates' visual cortex, was chosen as one of the saliency prediction models for its human-based inspiration. This model operates based on different stimuli and intensity feature maps, requiring no additional training \cite{itti}.

\textbf{DeepGaze IIE} -- The other saliency prediction model, employs deep learning techniques and is trained on a saliency dataset. It leverages existing neural networks pre-trained for object recognition tasks and is capable of handling multiple backbones for fixation prediction \cite{deepgaze}.

\begin{algorithm}[htbp]
\caption{Experimental Pipeline}\label{alg:experiment_pipeline}
\small
\begin{algorithmic}
\STATE \textbf{Input:} Dataset annotations, Prediction Model
\STATE Load annotations from the Dataset
\STATE Preprocess annotations
\STATE Load Prediction Model
\FOR{each image $i$ in the dataset}
    \STATE $p \gets \text{inferPrediction}(i)$
    \STATE $r \gets \text{computePearsonCorrelation}(p, \text{ground truth})$
\ENDFOR
\STATE $M \gets \text{meanCorrelationPerCategory}(\text{all images})$
\STATE \textbf{Output:} $M$
\end{algorithmic}
\end{algorithm}

\begin{table*}[t]
\centering
\renewcommand{\arraystretch}{1.15} 
\begin{tabularx}{\textwidth}{|>{\arraybackslash}p{4cm}|*{4}{>{\centering\arraybackslash}X|}c|}
\hline
\multicolumn{1}{|c|}{\multirow{2}{*}{\textbf{Technique}}} & \multicolumn{2}{c|}{\textbf{Mean Avg. Pearson Corr. (mA$\rho$)}} & \multicolumn{2}{c|}{\textbf{Avg. Runtime/image (s)}} & \multirow{2}{*}{\textbf{Model Type}} \\ \cline{2-5}
\multicolumn{1}{|c|}{} & \textbf{Pascal VOC} & \textbf{COCO} & \textbf{Pascal VOC} & \textbf{COCO} &  \\ \hline
Depth Anything & 0.273 & 0.125 & \textbf{0.020} & \textbf{0.029} & Depth Prediction \\ \hline
DPT-Large & 0.283 & 0.129 & 0.046 & 0.050 & Depth Prediction \\ \hline
Itti-Koch Model & 0.280 & 0.130 & 0.030 & 0.065 & Saliency Prediction \\ \hline 
DeepGaze IIE & \textbf{0.459} & \textbf{0.170} & 0.042 & 0.084  & Saliency Prediction \\ \hline \hline
\textit{Average} & \textit{0.324} & \textit{0.139} & \textit{0.035} & \textit{0.505} & \textit{N/A} \\ \hline
\end{tabularx}

\medskip

\caption{Evaluation results of various Depth and Saliency Prediction techniques on the Pascal VOC and COCO datasets with the respective metrics and their performance. The best-performing results are denoted in bold.}
\label{tab:results:single}
\end{table*}

\subsection{Experimental Setup}%
\label{sec:setup}

The algorithm outlined in Algorithm \ref{alg:experiment_pipeline} describes the experimental pipeline for evaluating the performance of the chosen prediction models using a given dataset. The process begins by loading the dataset annotations and the prediction model. For each image in the dataset, the model generates predictions, which are then compared to the ground truth using Pearson correlation (refer to Equation \ref{eq:pearson_correlation}). This correlation is computed for each image, and the mean correlation per category is calculated across all images (refer to Equation \ref{eq:mapc}).

\begin{figure}[ht]
    \centering
    \includegraphics[width=\columnwidth]{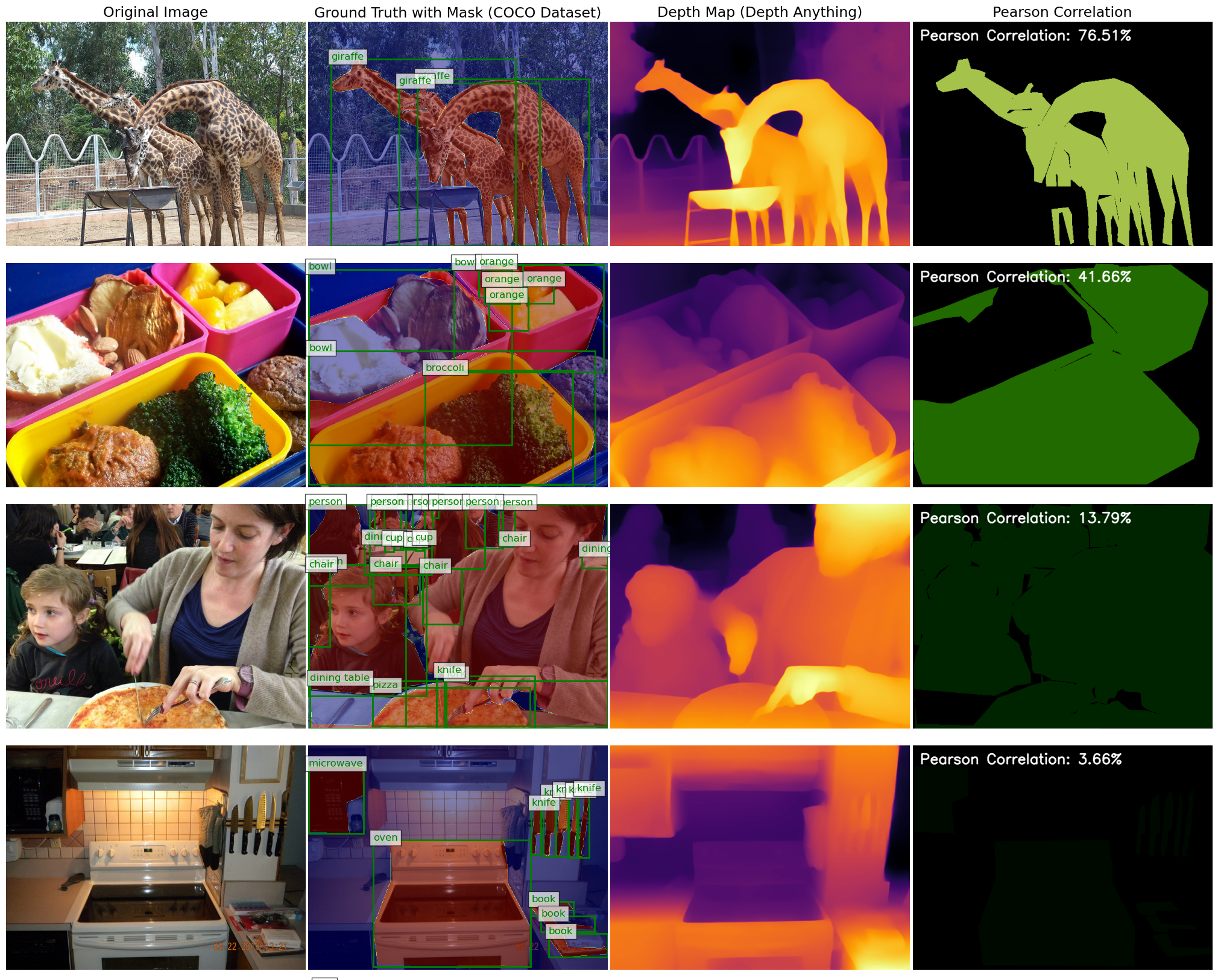}
    \caption{Sample images from the COCO dataset along with their corresponding ground truth masks, depth maps generated by the Depth Anything Model, and Pearson correlation values.}
    \label{fig:image_grid_DA}
\end{figure}

\begin{figure}[ht]
    \centering
    \includegraphics[width=\columnwidth]{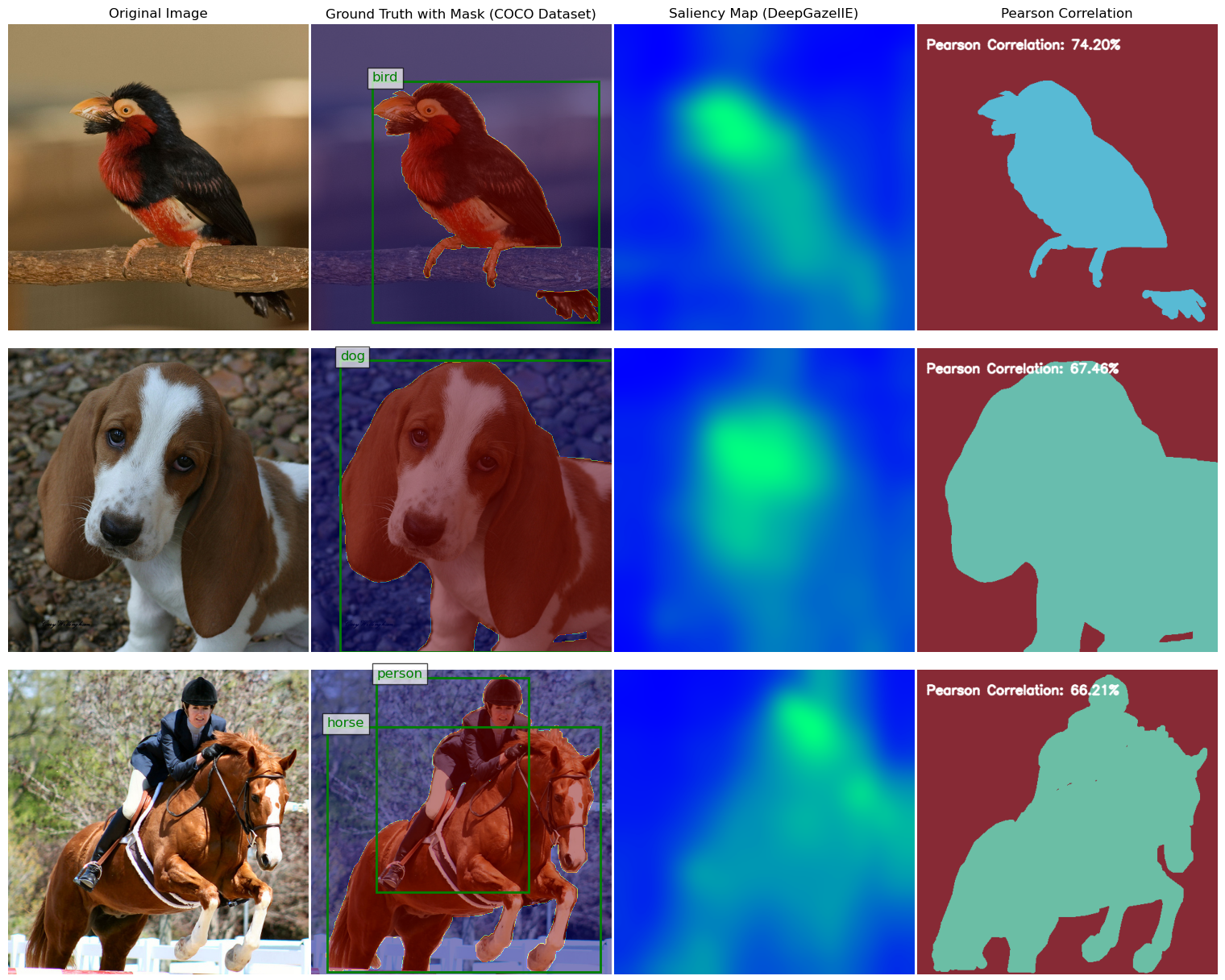}
    \caption{Sample images from the Pascal VOC dataset along with their corresponding ground truth masks, saliency maps generated by the DeepGaze IIE Model, and Pearson correlation values.}
    \label{fig:image_grid_DeepGaze}
\end{figure}

\section{Evaluation}%
\label{chp:evaluation}

\subsection{Metrics}%
\label{sec:metrics}

The primary evaluation metric used was Pearson correlation ($\rho$), as shown in Equation \ref{eq:pearson_correlation}. Pearson correlation measures the linear relationship between the ground truth and the generated depth or saliency map, evaluating how well the predictions align with the actual values, as described in \cite{pearson}. This metric focuses on the strength and direction of the correlation between two datasets, disregarding differences in intensity or scale. A higher Pearson correlation value indicates a stronger relationship between the predicted and true values, with a value of 1 representing a perfect linear correlation.

\begin{equation}
\rho_{X,Y} = \frac{\text{Cov}(X, Y)}{\sigma_X \sigma_Y}
\label{eq:pearson_correlation}
\end{equation}

To evaluate the overall performance across multiple classes or categories, we utilised the mean Average Pearson Correlation (mA$\rho$), which averages the Pearson correlation across all classes, as defined in Equation \ref{eq:mapc}. This metric provides a more comprehensive view of the model's performance by calculating the average Pearson correlation across all classes, where \( C \) represents the total number of classes or categories. For each class \( c \), the covariance between the ground truth and the predictions is divided by the product of their standard deviations. A higher mA$\rho$ value indicates stronger correlations across all classes, with a value of 1 signifying perfect alignment between predictions and ground truth for every class.

\begin{equation} \text{mA}\rho_{X,Y} = \frac{1}{C} \sum_{c=1}^{C} \frac{\text{Cov}(X_c, Y_c)}{\sigma_{X_c} \sigma_{Y_c}} 
\label{eq:mapc}
\end{equation}

\subsection{Results}%
\label{sec:results}

\subsubsection{Comparative Performance}
\label{subsec:comparison}

The proposed experiment was conducted using an NVIDIA GeForce RTX 4070 GPU. The Mean Average Pearson Correlation (mA$\rho$) and the average runtime per image were measured for each experiment on the Pascal VOC and COCO datasets, as presented in Table \ref{tab:correlation}. Additionally, Table \ref{tab:results:single} displays the Average Pearson Correlation Results per Class for various categories in the COCO dataset using the Depth Anything model. This table was included to assess the distribution of individual category results for the least-performing model.

\subsubsection{Dataset Analysis}
\label{subsec:analysis}

Comparing the results in Table \ref{tab:results:single}, DeepGaze IIE emerges as the most effective model, outperforming others with consistently higher Mean Average Pearson Correlation (mA$\rho$) values. However, the overall low mA$\rho$ scores on the COCO dataset, compared to the relatively better performance on PASCAL VOC, suggest that contextual complexity may challenge model accuracy. This discrepancy could be due to differences in dataset characteristics: PASCAL VOC’s smaller dataset size and larger object scales may provide less ambiguous visual contexts, allowing models to more easily detect and predict objects accurately. COCO, by contrast, focuses on non-iconic, complex scenes where objects frequently appear in diverse contexts with more categories and instances per image (averaging 3.5 categories and 7.7 instances per image), adding a layer of difficulty for models that may struggle to isolate relevant visual cues from background information \cite{coco_compare}. In COCO, only 10\% of images contain a single category, whereas over 60\% of PASCAL VOC images feature one category, further underscoring how COCO’s dense object presence and multi-object contexts might inhibit model precision \cite{coco_compare}.

Interestingly, DPT-Large, despite slightly slower execution times, achieved higher mA$\rho$ than Depth Anything, which was the fastest model, while the Itti-Koch model, inspired by the primate visual cortex, lagged in accuracy, highlighting the advantage of more sophisticated saliency algorithms. The stronger performance of DeepGaze IIE and DPT-Large suggests that visual saliency, rather than depth prediction alone, may contribute more significantly to detection accuracy, especially within diverse visual contexts. Individual class results, particularly for the least performing depth model (Table \ref{tab:correlation}), reveal that object size impacts correlation scores; larger objects, like animals and vehicles, demonstrated higher scores than smaller objects. Additionally, object placement and background context proved influential, as objects commonly appearing in the background, such as TVs, showed lower correlation scores, underscoring that object size, position, and scene complexity critically shape detection model effectiveness. These findings imply that models prioritising visual saliency could achieve better alignment with real-world object detection needs, especially in datasets with complex, non-iconic scenes.

\begin{table*}[t]
\centering
\footnotesize
\resizebox{\textwidth}{!}{ 
\begin{tabular}{|l|l|l|l|l|l|l|l|l|l|}
\hline
\rowcolor{headercolor}
\textcolor{textcolor}{\textbf{Category Name}} & \textcolor{textcolor}{giraffe} & \textcolor{textcolor}{airplane} & \textcolor{textcolor}{parking meter} & \textcolor{textcolor}{elephant} & \textcolor{textcolor}{horse} & \textcolor{textcolor}{stop sign} & \textcolor{textcolor}{zebra} & \textcolor{textcolor}{bear} & \textcolor{textcolor}{bed} \\
\hline
\rowcolor{rowcolor}
Avg. Corr. per Class & 0.326 & 0.310 & 0.304 & 0.301 & 0.291 & 0.288 & 0.280 & 0.279 & 0.259 \\
\hline
\rowcolor{headercolor}
\textcolor{textcolor}{Category Name} & \textcolor{textcolor}{cake} & \textcolor{textcolor}{motorcycle} & \textcolor{textcolor}{teddy bear} & \textcolor{textcolor}{hot dog} & \textcolor{textcolor}{person} & \textcolor{textcolor}{fire hydrant} & \textcolor{textcolor}{sandwich} & \textcolor{textcolor}{cat} & \textcolor{textcolor}{pizza} \\
\hline
\rowcolor{rowcolor}
Avg. Corr. per Class & 0.256 & 0.254 & 0.243 & 0.243 & 0.241 & 0.237 & 0.232 & 0.230 & 0.228 \\
\hline
\rowcolor{headercolor}
\textcolor{textcolor}{Category Name} & \textcolor{textcolor}{cow} & \textcolor{textcolor}{dining table} & \textcolor{textcolor}{suitcase} & \textcolor{textcolor}{dog} & \textcolor{textcolor}{donut} & \textcolor{textcolor}{sheep} & \textcolor{textcolor}{banana} & \textcolor{textcolor}{bird} & \textcolor{textcolor}{snowboard} \\
\hline
\rowcolor{rowcolor}
Avg. Corr. per Class & 0.224 & 0.221 & 0.221 & 0.206 & 0.203 & 0.196 & 0.184 & 0.153 & 0.122 \\
\hline
\rowcolor{headercolor}
\textcolor{textcolor}{Category Name} & \textcolor{textcolor}{laptop} & \textcolor{textcolor}{orange} & \textcolor{textcolor}{keyboard} & \textcolor{textcolor}{apple} & \textcolor{textcolor}{surfboard} & \textcolor{textcolor}{bus} & \textcolor{textcolor}{umbrella} & \textcolor{textcolor}{train} & \textcolor{textcolor}{kite} \\
\hline
\rowcolor{rowcolor}
Avg. Corr. per Class & 0.121 & 0.120 & 0.119 & 0.118 & 0.117 & 0.110 & 0.110 & 0.105 & 0.103 \\
\hline
\rowcolor{headercolor}
\textcolor{textcolor}{Category Name} & \textcolor{textcolor}{broccoli} & \textcolor{textcolor}{couch} & \textcolor{textcolor}{vase} & \textcolor{textcolor}{bowl} & \textcolor{textcolor}{skateboard} & \textcolor{textcolor}{boat} & \textcolor{textcolor}{tennis racket} & \textcolor{textcolor}{scissors} & \textcolor{textcolor}{frisbee} \\
\hline
\rowcolor{rowcolor}
Avg. Corr. per Class & 0.099 & 0.097 & 0.095 & 0.092 & 0.089 & 0.086 & 0.083 & 0.079 & 0.077 \\
\hline
\rowcolor{headercolor}
\textcolor{textcolor}{Category Name} & \textcolor{textcolor}{remote} & \textcolor{textcolor}{carrot} & \textcolor{textcolor}{bench} & \textcolor{textcolor}{tie} & \textcolor{textcolor}{traffic light} & \textcolor{textcolor}{cell phone} & \textcolor{textcolor}{bicycle} & \textcolor{textcolor}{skis} & \textcolor{textcolor}{toothbrush} \\
\hline
\rowcolor{rowcolor}
Avg. Corr. per Class & 0.076 & 0.075 & 0.074 & 0.074 & 0.073 & 0.071 & 0.070 & 0.070 & 0.070 \\
\hline
\rowcolor{headercolor}
\textcolor{textcolor}{Category Name} & \textcolor{textcolor}{hair drier} & \textcolor{textcolor}{backpack} & \textcolor{textcolor}{baseball glove} & \textcolor{textcolor}{sink} & \textcolor{textcolor}{truck} & \textcolor{textcolor}{mouse} & \textcolor{textcolor}{handbag} & \textcolor{textcolor}{clock} & \textcolor{textcolor}{toilet} \\
\hline
\rowcolor{rowcolor}
Avg. Corr. per Class & 0.065 & 0.064 & 0.061 & 0.060 & 0.054 & 0.054 & 0.053 & 0.043 & 0.042 \\
\hline
\rowcolor{headercolor}
\textcolor{textcolor}{Category Name} & \textcolor{textcolor}{baseball bat} & \textcolor{textcolor}{refrigerator} & \textcolor{textcolor}{sports ball} & \textcolor{textcolor}{knife} & \textcolor{textcolor}{oven} & \textcolor{textcolor}{potted plant} & \textcolor{textcolor}{chair} & \textcolor{textcolor}{fork} & \textcolor{textcolor}{cup} \\
\hline
\rowcolor{rowcolor}
Avg. Corr. per Class & 0.037 & 0.033 & 0.020 & 0.016 & 0.016 & 0.016 & 0.013 & 0.012 & 0.012 \\
\hline
\rowcolor{headercolor}
\textcolor{textcolor}{Category Name} & \textcolor{textcolor}{book} & \textcolor{textcolor}{spoon} & \textcolor{textcolor}{microwave} & \textcolor{textcolor}{bottle} & \textcolor{textcolor}{wine glass} & \textcolor{textcolor}{car} & \textcolor{textcolor}{toaster} & \textcolor{textcolor}{tv} & \textcolor{textcolor}{\textbf{mA$\rho$}} \\
\hline
\rowcolor{rowcolor}
Avg. Corr. per Class & 0.011 & 0.010 & 0.001 & -0.000 & -0.004 & -0.010 & -0.016 & -0.034 & \textbf{0.125} \\
\hline
\end{tabular}
}
\caption{Average Pearson Correlation Results per Class for Different Categories in the COCO Dataset using Depth Anything Model.}
\label{tab:correlation}
\end{table*}



\subsection{Discussion}%
\label{sec:discuss}


\subsubsection{Implications of Findings}
\label{subsec:implications}

The implications of these findings suggest that the correlations observed between depth prediction, saliency prediction, and object detection accuracy could serve as a basis for advancing multitask learning frameworks within object detection, as well as improve computational efficiency. The notable relationship between saliency prediction and detection accuracy indicates that incorporating saliency-focused tasks may enhance object detection models by aligning them more closely with human perceptual tendencies, especially in visually complex environments. By leveraging the strengths of both depth and saliency predictions, object detection models could benefit from the contextual information provided by these tasks, potentially leading to improved robustness and accuracy in diverse and challenging scenes.

Furthermore, the results highlight that saliency prediction, demonstrating a notably higher correlation with detection accuracy than depth prediction, may serve as a valuable tool for evaluating and refining object detection datasets. By examining the specific saliency aspects that contribute most to detection accuracy, researchers and practitioners could assess dataset viability for certain detection tasks and incorporate saliency cues directly into detection architectures. Such integration could streamline multitask learning, enabling models to more effectively prioritise salient features and, thereby, enhance detection performance across varied image contexts.

In contrast, while depth prediction showed lower correlations with detection accuracy, it holds potential for tasks involving the assessment of object size and scale differentiation. Depth models, although less directly correlated with detection accuracy, can still play an essential role in identifying size distinctions between large and small objects across diverse backgrounds \cite{small_detection}, and distance scales, such as those found in the SODA dataset \cite{soda}.

\subsubsection{Limitations of Dataset Design}
\label{subsec:datasets}
While this paper primarily explored the correlation between depth estimation and saliency prediction in relation to object detection, it also highlighted the critical importance of thoughtful dataset design, as evidenced by the discrepancy in results between the Pascal VOC and COCO datasets. A balanced dataset that adequately represents various object sizes, scales, and categories is essential for training robust object detection models. The findings highlight that the average number of categories and instances per image can significantly influence model performance. Specifically, the integration of diverse object representations and a careful consideration of how these elements interact within the dataset can enhance the overall effectiveness of detection algorithms, leading to improved accuracy and reliability in real-world applications.

Furthermore, the models utilised in this study, particularly those focused on saliency prediction, exhibit a notable emphasis on centre bias \cite{sara}, which reflects the subjectivity inherent in saliency estimation. This subjectivity arises from the tendency of models to prioritise central objects within an image, potentially overlooking important contextual elements \cite{sara}. In contrast, the COCO dataset is characterised by its inclusion of non-iconic scenes that present objects within complex and varied backgrounds. This focus on realistic scenarios underscores the necessity of developing models that can effectively navigate and interpret such contexts, where the traditional assumptions of saliency may not apply. 

\section{Conclusion}%
\label{chp:conclusion}

This paper investigates the relationships between object detection accuracy and two essential visual tasks: depth prediction and visual saliency prediction. Our experiments with state-of-the-art models (DeepGaze IIE, Depth Anything, DPT-Large, and Itti's model) on COCO and Pascal VOC datasets reveal that visual saliency demonstrates consistently stronger correlations with detection accuracy, achieving a Mean Average Pearson Correlation (mA$\rho$) up to 0.459 on Pascal VOC, compared to depth prediction (mA$\rho$ up to 0.283). Larger objects show correlation values up to three times higher than smaller ones, highlighting the impact of object scale on model performance. These findings suggest that incorporating visual saliency features into object detection frameworks could be particularly beneficial for specific object categories. Moreover, the observed category-specific variations offer valuable insights for optimising feature engineering and guiding dataset design, potentially leading to more efficient and accurate object detection systems aligned with human object perception.


\bibliographystyle{IEEEtran}
\bibliography{references}

\begin{thebibliography}{10}
\providecommand{\url}[1]{#1}
\csname url@samestyle\endcsname
\providecommand{\newblock}{\relax}
\providecommand{\bibinfo}[2]{#2}
\providecommand{\BIBentrySTDinterwordspacing}{\spaceskip=0pt\relax}
\providecommand{\BIBentryALTinterwordstretchfactor}{4}
\providecommand{\BIBentryALTinterwordspacing}{\spaceskip=\fontdimen2\font plus
\BIBentryALTinterwordstretchfactor\fontdimen3\font minus \fontdimen4\font\relax}
\providecommand{\BIBforeignlanguage}[2]{{%
\expandafter\ifx\csname l@#1\endcsname\relax
\typeout{** WARNING: IEEEtran.bst: No hyphenation pattern has been}%
\typeout{** loaded for the language `#1'. Using the pattern for}%
\typeout{** the default language instead.}%
\else
\language=\csname l@#1\endcsname
\fi
#2}}
\providecommand{\BIBdecl}{\relax}
\BIBdecl

\bibitem{human_perception}
L.~Fei-Fei, A.~Iyer, C.~Koch, and P.~Perona, ``What do we perceive in a glance of a real-world scene?'' \emph{Journal of Vision}, vol.~7, no.~1, p.~10, 2007.

\bibitem{bartolo2024integrating}
M.~Bartolo, D.~Seychell, and J.~Bajada, ``Integrating saliency ranking and reinforcement learning for enhanced object detection,'' \emph{arXiv preprint arXiv:2408.06803}, 2024.

\bibitem{perception}
J.~Johnson, ``Designing with the mind in mind: The psychological basis of user interface design guidelines,'' in \emph{Extended Abstracts of the 2021 CHI Conference on Human Factors in Computing Systems}, ser. CHI EA '21.\hskip 1em plus 0.5em minus 0.4em\relax New York, NY, USA: Association for Computing Machinery, 2021.

\bibitem{vandenhende2022multitasklearningvisualscene}
S.~Vandenhende, ``Multi-task learning for visual scene understanding,'' 2022.

\bibitem{Pisani2024DetectingLF}
D.~Pisani and D.~Seychell, ``Detecting litter from aerial imagery using the soda dataset,'' \emph{2024 IEEE 22nd Mediterranean Electrotechnical Conference (MELECON)}, pp. 897--902, 2024.

\bibitem{small_detection}
M.~Schembri and D.~Seychell, ``Small object detection in highly variable backgrounds,'' in \emph{2019 11th International Symposium on Image and Signal Processing and Analysis (ISPA)}, 2019, pp. 32--37.

\bibitem{yolov10}
\BIBentryALTinterwordspacing
A.~Wang, H.~Chen, L.~Liu, K.~Chen, Z.~Lin, J.~Han, and G.~Ding, ``{YOLOv10}: Real-time end-to-end object detection,'' 2024. [Online]. Available: \url{https://arxiv.org/abs/2405.14458}
\BIBentrySTDinterwordspacing

\bibitem{yolo11}
\BIBentryALTinterwordspacing
G.~Jocher and J.~Qiu, ``Ultralytics {YOLO11},'' 2024. [Online]. Available: \url{https://github.com/ultralytics/ultralytics}
\BIBentrySTDinterwordspacing

\bibitem{rtdetr}
Y.~Zhao, W.~Lv, S.~Xu, J.~Wei, G.~Wang, Q.~Dang, Y.~Liu, and J.~Chen, ``{DETRs} beat {YOLOs} on real-time object detection,'' 2024.

\bibitem{depth_estimation}
I.~Vasiljevic, N.~I. Kolkin, S.~Zhang, R.~Luo, H.~Wang, F.~Z. Dai, A.~F. Daniele, M.~Mostajabi, S.~Basart, M.~R. Walter, and G.~Shakhnarovich, ``{DIODE:} {A} dense indoor and outdoor depth dataset,'' \emph{CoRR}, vol. abs/1908.00463, 2019.

\bibitem{depth_anything}
L.~Yang, B.~Kang, Z.~Huang, X.~Xu, J.~Feng, and H.~Zhao, ``Depth anything: Unleashing the power of large-scale unlabeled data,'' 2024.

\bibitem{itti}
L.~Itti, C.~Koch, and E.~Niebur, ``A model of saliency-based visual attention for rapid scene analysis,'' \emph{IEEE Transactions on Pattern Analysis and Machine Intelligence}, vol.~20, no.~11, pp. 1254--1259, 11 1998.

\bibitem{deepgaze}
A.~Linardos, M.~K{\"{u}}mmerer, O.~Press, and M.~Bethge, ``Calibrated prediction in and out-of-domain for state-of-the-art saliency modeling,'' \emph{CoRR}, vol. abs/2105.12441, 2021.

\bibitem{sara}
D.~Seychell and C.~J. Debono, ``Ranking regions of visual saliency in rgb-d content,'' in \emph{2018 International Conference on 3D Immersion (IC3D)}, 2018, pp. 1--8.

\bibitem{object_detection_philospy}
T.~Boger and T.~Ullman, ``What is "where": Physical reasoning informs object location,'' \emph{Open Mind (Cambridge)}, vol.~7, pp. 130--140, 5 2023.

\bibitem{cocodataset}
T.~Lin, M.~Maire, S.~J. Belongie, L.~D. Bourdev, R.~B. Girshick, J.~Hays, P.~Perona, D.~Ramanan, P.~Doll{'{a} }r, and C.~L. Zitnick, ``Microsoft {COCO:} common objects in context,'' \emph{CoRR}, vol. abs/1405.0312, 2014.

\bibitem{pascal_voc_2012}
M.~Everingham, L.~Van~Gool, C.~K.~I. Williams, J.~Winn, and A.~Zisserman, ``The {PASCAL} {V}isual {O}bject {C}lasses {C}hallenge 2012 {(VOC2012)} {R}esults,'' http://www.pascal-network.org/challenges/VOC/voc2012/workshop/index.html.

\bibitem{dpt_large}
R.~Ranftl, A.~Bochkovskiy, and V.~Koltun, ``Vision transformers for dense prediction,'' \emph{CoRR}, vol. abs/2103.13413, 2021.

\bibitem{pearson}
P.~Sedgwick, ``Pearson's correlation coefficient,'' \emph{BMJ}, vol. 345, pp. e4483--e4483, 07 2012.

\bibitem{coco_compare}
W.~Zhang, ``A fruit ripeness detection method using adapted deep learning-based approach,'' \emph{International Journal of Advanced Computer Science and Applications}, vol.~14, 01 2023.

\bibitem{soda}
D.~Pisani, D.~Seychell, C.~J. Debono, and M.~Schembri, ``Soda: A dataset for small object detection in uav captured imagery,'' in \emph{2024 IEEE International Conference on Image Processing (ICIP)}, 2024, pp. 151--157.

\end{thebibliography}

\end{document}